\documentclass{article}

     \usepackage[final]{neurips_2020}

\usepackage[utf8]{inputenc} 
\usepackage[T1]{fontenc}    
\usepackage{hyperref}       
\usepackage{url}            
\usepackage{booktabs}       
\usepackage{amsfonts}       
\usepackage{nicefrac}       
\usepackage{microtype}      
\usepackage{times}
\usepackage{epsfig}
\usepackage{graphicx}
\usepackage{amsmath}
\usepackage{amssymb}
\usepackage{graphicx}
\usepackage{natbib}
\usepackage{subcaption} 
\usepackage[export]{adjustbox}
\title{Do Vision-Language Foundational models show Robust Visual Perception?}

\author{
Shivam Chandhok$^{1,2}$ \quad Pranav Tandon$^{1}$\vspace{2mm} \\
$^1$University of British Columbia \quad $^2$Vector Institute for AI \vspace{1mm} \\
\texttt{\small \{chshivam\}@cs.ubc.ca}
}

\begin{document}

\maketitle

\begin{abstract}
  Recent advances in vision-language
foundational models have enabled development of systems that can perform visual understanding and reasoning tasks.
However, it is unclear if these models are robust to distribution shifts, and how their performance and generalization capabilities vary under changes in data distribution. In this project we strive to answer the question "Are vision-language foundational models robust to distribution shifts like human perception?" Specifically,   we consider a diverse range of vision-language models and compare how the performance of these systems is affected by corruption based distribution shifts (such as \textit{motion blur, fog, snow, gaussian noise})  commonly found in practical real-world scenarios. We analyse the generalization capabilities qualitatively and quantitatively on zero-shot image classification task under aforementioned distribution shifts.
Our code will be avaible at \url{https://github.com/shivam-chandhok/CPSC-540-Project}
\end{abstract}

\section{Introduction}

Advances in deep learning have led to the development of models that can perform visual understanding and reasoning tasks with performance at par with human visual perception \cite{clip, resnet, vit, llava}. Majority of these systems follow the supervised learning paradigm where the model is trained on a dataset with pairs of images and corresponding annotations \cite{resnet,vit}. However, the supervised learning paradigm relies on the \underline{assumption} that the data at test-time follows the same distribution of images that the model had seen during training. This underlying assumption inhibits the applicability of deep learning models in \underline{practical scenarios} where this \underline{assumption seldom holds}. In practical real-world problem settings, the distribution of images at test-time might be different than the ones during training and performance of supervised models \underline{drastically deteriorates under such circumstances}
\cite{pacs,corrupt,distritransformer,intriguing}.

This can be attributed to the \underline{dataset bias} problem \cite{pacs,corrupt,distritransformer} where deep learning models tend to develop a bias toward the data on which they are trained by overfitting to nuiances such as noise, appearance, background, color, texture.  For example, given a model trained on real-world photographs of cats and dogs, it finds it very hard to recognize sketches of cats (\textit{change in style} ) or cats in foggy or snowy backgrounds (\textit{change in appearance and background}). This is because  there is a \underline{distribution-shift} between real-world photographs and sketches or foggy images of cats and dogs \cite{corrupt}. 
Thus, practical applications demand the development of models that are robust to nuisances and can generalize to novel unseen views or depictions of an object.

Recently, the deep learning community has witnessed a boom in the development of large vision and language models which are also referred to as \underline{foundational models} \cite{clip,coca,llava,blip2}. These models are trained on large scale multimodal data from the web with a paradigm called \underline{Contrastive learning} which helps them learn generalizable representation. These models transfer well to various downstream image and video understanding tasks and generalize well to new classes in a zero-shot setting \cite{clip,coca}.

However, there is a lack of understanding and rigourous analysis on how these models perform under distribution-shifts. There is some evidence on how these models generalize to changes in style, (say sketches or cartoons of cats and dogs) \cite{clip} however this can be due to the fact that such images are commonly available on the web and these models were trained on them. It is important to analyse the capability of these models to tackle distribution shifts like common corruptions (\textit{motion blur, gaussian noise}) \cite{corrupt}that are often found in real-world scenarios but not too often found in images on the web which the models where trained on.
\vspace{-12pt}
\section{Contribution and Scope}
\vspace{-8pt}
To this end, in this work, we \underline{study the robustness} of various foundational models under varying levels of corruptions (i.e \textit{gaussian noise, fog, snow, motion blur}) \cite{corrupt} which are commonly found in practical real world scenarios. Specifically, we \underline{first categorize} the different types of foundational models based on the \underline{objectives they are trained to optimize} and then compare (qualitatively and quantitatively) the variation in \underline{performance and latent space representations} with varying levels of diverse corruptions.

Foundational models are widely being used in variety of real-world applications like autonomous driving, robotics, medical imaging. Analysing their performance under diverse distribution shifts is an important contribution to the community as this can help users make informed choices on which models to use for a given application. For e.g it will be useful for \underline{autonomous driving engineers} to understand which models are robust to \underline{weather effects like fog and snow and blurring effects like motion blur} which are commonly encountered while driving. This will help them select models which are robust to such variations for object recognition and detection in autonomous driving systems. Furthermore, \underline{medical experts} using foundational models can benefit from analysis on noise based perturbations as \underline{MRI or CT-Scan images contain specific noise perturbations} due to type of machines and setup used to acquire such scans.
\vspace{-12pt}
\section{Literature Review}
\vspace{-10pt}
\label{litrev}
In this section, we first categorize the image understanding models (that we will study) based on the objective they optimize, scale and type of data they are trained on and architecture. Broadly there are two groups 1) Supervised Vision Models 2) Multimodal Foundational Models.
Additionally, we further categorize foundational models into 1) Contrastive Multi-Encoder models 2) Encoder-Decoder Generative models
3) Hybrid-Models

\textbf{Supervised Vision Models}: These models are commonly trained with image-annotation paired data which is specifically tailored for a given downstream task at hand (For e.g, ResNet-50 trained ImageNet object classification \cite{resnet,vit}). These models  are usually trained with moderate scale benchmark datasets and optimize some form of cross-entropy objective between input images and labels. They are commonly single-encoder models that can be seen as specialists at the task they have been trained to perform and struggle to generalize to new tasks or data distributions at test-time \cite{intriguing, resnet}. 


\textbf{Foundational Models}: Multimodal models like CLIP \cite{clip}, also referred to as vision-language foundational models can be seen as generalist models that aim to learn useful representation which can transfer to different use-cases, data distributions and downstream tasks.  These models are usually multimodal and learn from weakly aligned image-text data readily available on the web. This drastically reduces the annotation cost  since we do not need human-annotations to accurately give labels to image examples \cite{coca,clip}.

Furthermore, these models are commonly trained on large-scale data using some form of \underline{Contrastive learning objective}. Contrastive learning is a paradigm \cite{clip} that aims to project high dimensional data like images and text into a low dimensional manifold or embedding space where semantically similar inputs cluster together and lie in the same region and dissimilar inputs lie far apart. This helps them learn generalizable semantically rich representations which transfer well to novel tasks and distributions.\\
\vspace{-20pt}
\paragraph{Catagorization of Foundational Models}:The vision-language foundational models can be broadly categorized into 3 groups based on the specific objective they optimize and architecture they use. 

\textbf{1) Contrastive Multi-Encoder models} : These models employ a separate encoder for each modality (i.e image and text) and use some form of contrastive cross-modal alignment loss which ensures that similar inputs from both modalities lie close to each other (and dissimlar inputs farther apart) in a low dimensional embedding space. They learn separate image and text representations which are aligned and can be used for tasks like classification; but do not have joint components to learn fused image and text representations (required for complex tasks like visual question answering).For example the CLIP model \cite{clip} falls in this group of foundational models.

\textbf{2) Encoder-Decoder Generative models}: These models employ an encoder to project images to a low-dimensional representation and then use it as context to generate language descriptions or captions using a language decoder model. They are trained on weakly-aligned image caption pairs with language modeling loss (or captioning loss) on predicted and ground truth captions. They learn joint image and text representations for multimodal understanding tasks
that can be used for complex tasks like visual question answering; but do not produce text-only representations aligned with image embeddings, thus being less effective for zero-shot classification or retrieval tasks.

\textbf{3) Hybrid-Models}: These models aim to combine the best of both worlds and train jointly to optimize both contrastive loss and captioning loss. This helps them to build global representation which are aligned across modalities like Contrastive Multi-Encoder models and also learn joint image-text fused representations with language loss to understand fine-grained details about images and handle multimodal understanding tasks like visual question answering. \cite{blip2, coca}
\begin{figure*}
  \centering
  \includegraphics[width=\textwidth]{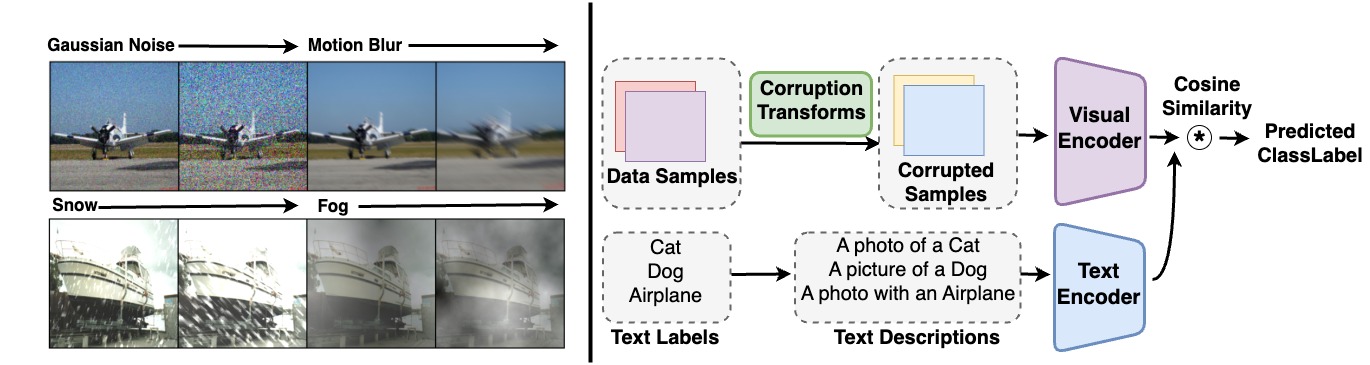} 
  \caption{\textbf{Left}: Example images with common corruptions (\textit{gaussian noise, motion blur, snow, fog}) with varying levels of severity (severity increases left to right). \textbf{Right}: Overall pipeline for our robustness analysis. Data samples (i.e images) are perturbed with common corruptions and passed through the visual encoder of the foundational model. Finally class label is predicted based on cosine similarity with textual encodings of class descriptions. }
  \vspace{-16pt}
  \label{fig:myfigure}
\end{figure*}
\vspace{-12pt}
\section{Method}
\vspace{-10pt}
\label{gen_inst}
\paragraph{Method Overview}: Our aim is to analyse the robustness of foundational vision-langiage models to common corruptions and study the variation in their performance and generalization. To this end, we study zero-shot classification performance of these models to validate their robustness. Figure \ref{fig:myfigure} shows the overall pipeline for our approach.
On the left, we show common corruption transformations we use to corrupt our input images. These include \textit{gaussian noise, motion blur, snow and fog} effects \cite{corrupt}.

On the right, we show the overview of our pipeline. We take datasamples from a supervised dataset and perturb them with common corruption transformations. These corrupted samples are passed through the pre-trained visual encoder branch (frozen) of a foundational model which outputs features in a low-dimensional embedding space. Finally, we compute the similarity between the image feature vector and textual encodings of class descriptions and convert the scores to class probabilities.
\vspace{-12pt}
\paragraph{Datasets and Evaluation protocol}: For our robustness study we use the CIFAR10 dataset which has 32x32 RGB images distributed over 10 classes (evaluate on the test set with 10,000). Furthermore, in order to evaluate on high-resolution real-world images, we include the PASCAL VOC dataset with 20 classes (evaluate on test set images with single-class objects ). We report the top-1 zero-shot classification accuracy for all cases
\vspace{-12pt}
\paragraph{Selected Models}: We study a variety of important foundational models and analyse variation in performance with corruptions. This includes the ResNet and Vision transformer variants of CLIP model \cite{clip} which fall in the category of Contrastive Multi-Encoder models (as discussed in section \ref{litrev}). We study the CoCa model \cite{coca} which uses a combination of contrastive loss and captioning (generative) loss and falls in the category of Hybrid models (as discussed in section \ref{litrev}). Finally, we study a recent image captioning model BLIP2 \cite{blip2} which combines image-text similarity matching, image-text contrastive objective and captioning language loss to build a strong hybrid foundational model.
\section{Results and Analysis}
In this section we study the robustness of selected foundational models with qualitative and quantitative results
on common corruptions. Table 1, 2, 3 and 4 show the quantitative results for \underline{\textit{Snow, Fog, Motion Blur and Gaussian Noise} corruption}, respectively with varying levels of severity on the CIFAR10 and PASCAL VOC dataset.
\vspace{-12pt}
\subsection{Analysis on CIFAR10 Dataset}
\vspace{-5pt}
\paragraph{Performance Without Corruptions} For performance on zero-shot classification without any corruptions (represented with column \textit{None} in Tables 1,2,3,4) we notice that the ResNet backbone based CLIP model performs considerably inferior to the other vision transformer based models (i.e CLIP ViT, CoCa, BLIP2).
This is in line with the observations in previous works \cite{intriguing, vit} where transformer based architectures perform much better than convolution based counterparts for object classification. Among the models with transfomer based backbone, we notice that CLIP and CoCa perform at par with each other but BLIP2 outperforms both of them in terms of performance. This can be attributed to the fact that BLIP2 is trained with a hybrid of 3 objectives (i.e \textit{image-text matching, contrastive and captioning loss}) which allow it to learn better more generalizable representation that transfer well to novel classes in zero-shot setting.
\paragraph{Performance With Corruptions}
We notice that in terms of average performance over severity of corruptions, the ResNet based CLIP model experiences considerable deterioration in performance compared to vision transformer based counterparts (i.e CLIP ViT, CoCa, BLIP2). This is in line with observations in previous works that find vision transformers to be much more robust to domain-shift than convolution based ResNet models \cite{intriguing}. We find that the Coca model is more robust than the CLIP ViT based model for most corruptions (all except for gaussian noise where it is slightly inferior) based on average accuracy. This trend can be explained with the fact that Coca is a hybrid model trained with both contrastive and captioning loss whereas CLIP is a multi-encoder model with only contrastive loss. Finally, we notice that BLIP2 has the best average score across severity of corruptions for all types of transformations. Thanks to the combination of \textit{image-text matching, contrastive and captioning loss}) it shows best robustness to common corruptions. 
Furthermore, we find that across corruptions the performance deteriorates the most for \textit{gaussian noise} and the least reduction of performance is for \textit{snow} perturbation.
\vspace{-10pt}
\subsection{Analysis on PASCAL VOC Dataset}
\vspace{-6pt}
\paragraph{Performance Without Corruptions}
Firstly, we notice that performance on PASCAL VOC is better than that on CIFAR10 which could be due to higher-resolution of PASCAL VOC images (which is close to the image resolution on which foundational models are originally trained) or smaller test-set size of PASCAL VOC.
For performance on zero-shot classification without any corruptions 
we notice a similar trend to CIFAR 10 case where that the ResNet backbone based CLIP model performs considerably inferior to the other vision transformer based models (i.e CLIP ViT, CoCa, BLIP2). We also notice that Coca hybrid model performs better than ViT based CLIP counterpart. Finally, we observe that unlike the CIFAR10 case, the BLIP2 model performs a little inferior to other models for PASCAL VOC. 
This could be due to the fact that images in PASCAL VOC are much more complex than images in CIFAR 10 and since BLIP2 is trained for being a good captioner, its performance on zero-shot classification reduces as image become more complex.
\paragraph{Performance With Corruptions}
Similar to the trend observed for CIFAR10, we notice that in terms of average performance over severity of corruptions, the ResNet based CLIP model experiences considerable deterioration in performance when compared to other vision transformer based models. Furthermore, comparing Coca model and ViT based CLIP model, we notice a similar trend as in CIFAR10 case where Coca hybrid model is more robust thanks to a combination of captioning and contrastive loss. Overall, both the hybrid models i.e CoCa and BLIP2 generalize considerable better under corruptions compare to multi-encoder contrastive models like CLIP.
Furthermore, we find that across corruptions the performance deteriorates the most for gaussian noise and the least reduction of performance is for fog perturbation.

\section{Qualitative Visualizations}
\vspace{-10pt}
In this section, we discuss qualitative visualization of latent space features for a \underline{Contrastive Multi-Encoder models (i.e CLIP ViT)} and a \underline{Hybrid Model (i.e CoCa}. 

Specifically, we visualize 100 randomly sampled 512- dimensional visual features for each class of CIFAR10 dataset under different types and levels of perturbation. For types of corruptions, we choose the \textit{Snow Corruption} which is one of the most common natural effect found in practical real-world scenarios, especially in Canada.

Figure 2 shows the latent space visualizations. Note that in case of No Corruptions applied to inputs, the different classes of CIFAR10 form 10 distinct clusters. The clusters are much more distinct and separated for the Hybrid CoCa (Figure 2, bottom row) vs CLIP ViT model (Figure 2, top row) showing that hybrid models trained with combination of contrastive and captioning loss learn better, more linearly separable and discriminative representations.

For snow corruption (severity=2), we notice that the clusters start to disappear much more for the CLIP ViT model than for the CoCa model where for we can still see distinct clusters for Coca . This shows that the Hybrid models are more robust to perturbations in its inputs.

As severity of corruption increases (i.e severity=5), we notice that both models end up with completely random points with no discernible clustering of points across classes. This trend aligns with the quantitative results where, as corruption severity is increased the zero-shot discriminative performance reduces as the latent space become less linearly separable.
\begin{table}[h!!]
  \centering
  \footnotesize 
  \begin{tabular}{|l|c|c|c|c|c|c|c|c|}
    \hline
    Models & \multicolumn{4}{c|}{CIFAR 10} & \multicolumn{4}{c|}{PASCAL VOC} \\
    \cline{2-9}
    & None & Severity=2 &  Severity=5 & Avg.&None & Severity=2 & Severity=5 & Avg. \\
    \hline
    CLIP (ResNet-101) &44.0 &22.0 &18.6 &28.2 &85.0 &63.0 &50.8 &66.2 \\
    CLIP (ViT-B) &93.8 &71.6 &53.1 &72.8 &94.2 &81.0 &72.6 &82.6\\
    CoCa &93.6 &76.8 &59.1 &76.5 &97.0 &91.2 &86.5 &\textbf{91.5}\\
    BLIP2 &99.0 &85.8 &69.2 &\textbf{84.7} &81.0 &78.5 &76.6 &79.0\\
    \hline
  \end{tabular}
  \caption{Quantitative results for \underline{\textit{Snow} Corruption} with varying levels of severity on CIFAR10 and PASCAL VOC datasets. We report the top-1 accuracy for zero-shot classification}
\end{table}

\vspace{-20pt}
\begin{table}[h!]
  \centering
  \footnotesize 
  \begin{tabular}{|l|c|c|c|c|c|c|c|c|}
    \hline
    Models & \multicolumn{4}{c|}{CIFAR 10} & \multicolumn{4}{c|}{PASCAL VOC} \\
    \cline{2-9}
    & None & Severity=2 &  Severity=5 & Avg.&None & Severity=2 & Severity=5 & Avg. \\
    \hline
    CLIP (ResNet-101) &44.0 &26.2 &11.8 &27.3 &85.0 &81.0 &73.1 &79.7 \\
    CLIP (ViT-B) &93.8 &76.0 &27.1 &65.6 &94.2 &92.7 &82.5 &89.8\\
    CoCa &93.6 &76.7 &28.0 &66.1 &97.0 &96.0 &91.5 &\textbf{94.8}\\
    BLIP2 &99.0 &86.0 &28.0 &\textbf{71.0} &81.0 &79.5 &79.0 &80.0\\
    \hline
  \end{tabular}
  \caption{Quantitative results for \underline{\textit{Fog} Corruption} with varying levels of severity on CIFAR10 and PASCAL VOC datasets. We report the top-1 accuracy for zero-shot classification}
\end{table}

\vspace{-20pt}
\begin{table}[h!!]
  \centering
  \footnotesize 
  \begin{tabular}{|l|c|c|c|c|c|c|c|c|}
    \hline
    Models & \multicolumn{4}{c|}{CIFAR 10} & \multicolumn{4}{c|}{PASCAL VOC} \\
    \cline{2-9}
    & None & Severity=2 &  Severity=5 & Avg.&None & Severity=2 & Severity=5 & Avg. \\
    \hline
    CLIP (ResNet-101) &44.0 &15.0 &14.2 &24.3 &85.0 &77.4 &52.0 &71.4 \\
    CLIP (ViT-B) &93.8 &43.0 &19.2 &52.0 &94.2 &88.3 &57.7 &80.0\\
    CoCa &93.6 &50.0 &21.2 &55.0 &97.0 &92.7 &68.2 &\textbf{86.0}\\
    BLIP2 &99.0 &68.0 &32.4 &\textbf{66.4} &81.0 &77.2 &75.0 &77.7\\
    \hline
  \end{tabular}
  \caption{Quantitative results for \underline{\textit{Motion Blur} Corruption} with varying levels of severity on CIFAR10 and PASCAL VOC datasets. We report the top-1 accuracy for zero-shot classification}
\end{table}

\vspace{-20pt}
\begin{table}[h!!]
  \centering
  \footnotesize 
  \begin{tabular}{|l|c|c|c|c|c|c|c|c|}
    \hline
    Models & \multicolumn{4}{c|}{CIFAR 10} & \multicolumn{4}{c|}{PASCAL VOC} \\
    \cline{2-9}
    & None & Severity=2 &  Severity=5 & Avg.&None & Severity=2 & Severity=5 & Avg. \\
    \hline
    CLIP (ResNet-101) &44.0 &12.4 &9.1 &21.8 &85.0 &60.3 &5.8 &50.3 \\
    CLIP (ViT-B) &93.8 &34.1 &14.2 &47.3 &94.2 &87.6 &36.3 &72.7\\
    CoCa &93.6 &48.6 &13.6 &52.0 &97.0 &92.6 &52.8 &\textbf{80.8}\\
    BLIP2 &99.0 &72.4 &18.5 &\textbf{63.3} &81.0 &76.5 &67.0 &74.8\\
    \hline
  \end{tabular}
  \caption{Quantitative results for \underline{\textit{Gaussian Noise} Corruption} with varying levels of severity on CIFAR10 and PASCAL VOC datasets. We report the top-1 accuracy for zero-shot classification}
\end{table}

\begin{figure*}
  \centering

  \begin{subfigure}{0.30\textwidth}
    \includegraphics[width=1.2\linewidth]{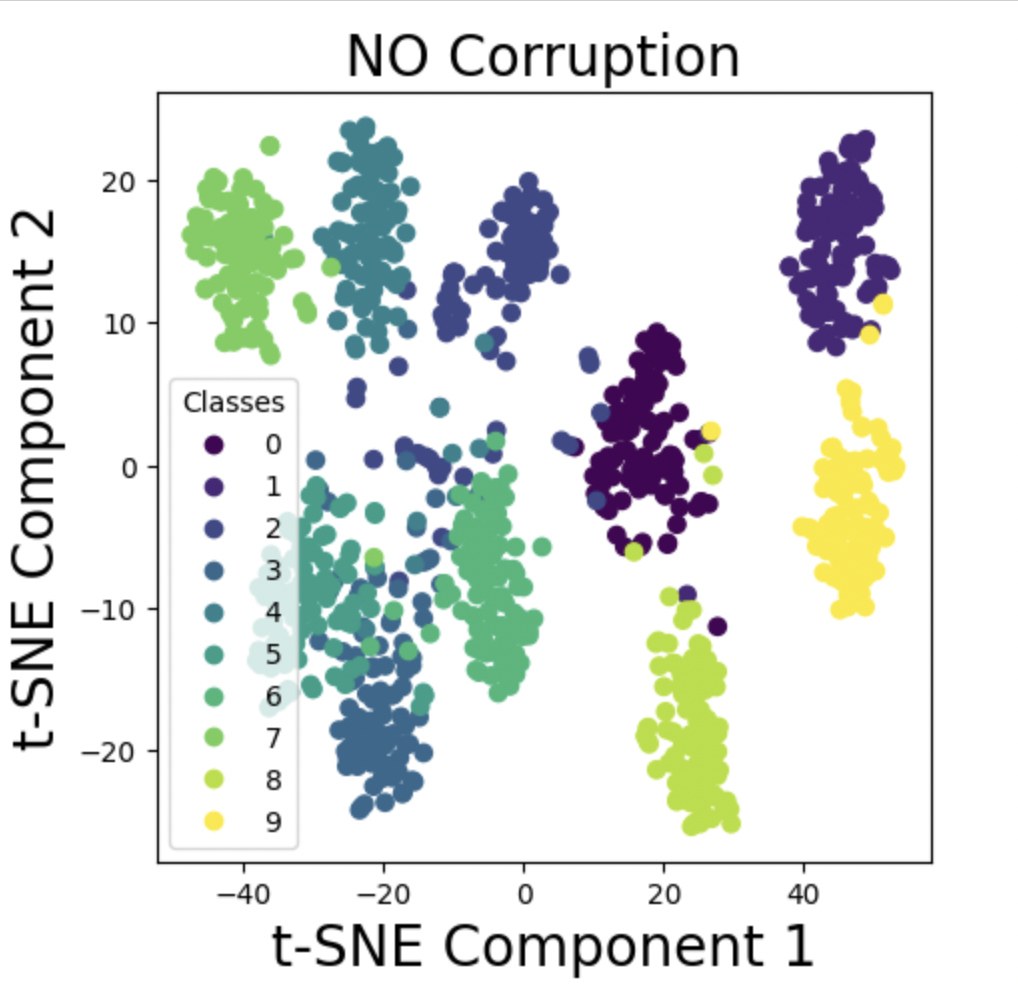} 
    \caption{Feature visualization for CLIP (ViT) with no corruption}
  \end{subfigure}
  \hfill
  \begin{subfigure}{0.30\textwidth}

    \includegraphics[width=1.2\linewidth]{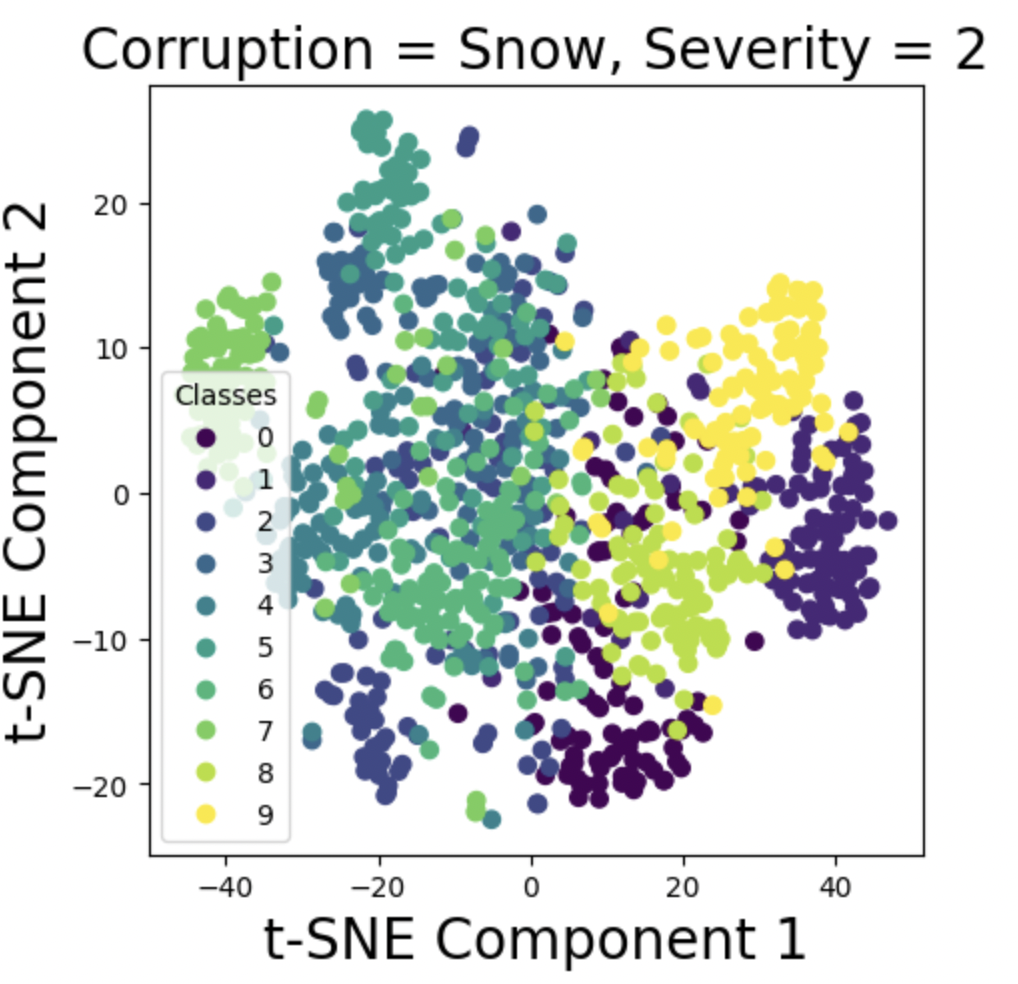} 
    \caption{Feature visualization for CLIP (ViT) with Snow (severity=2)}
  \end{subfigure}
  \hfill
  \begin{subfigure}{0.30\textwidth}
  
    \includegraphics[width=1.2\linewidth]{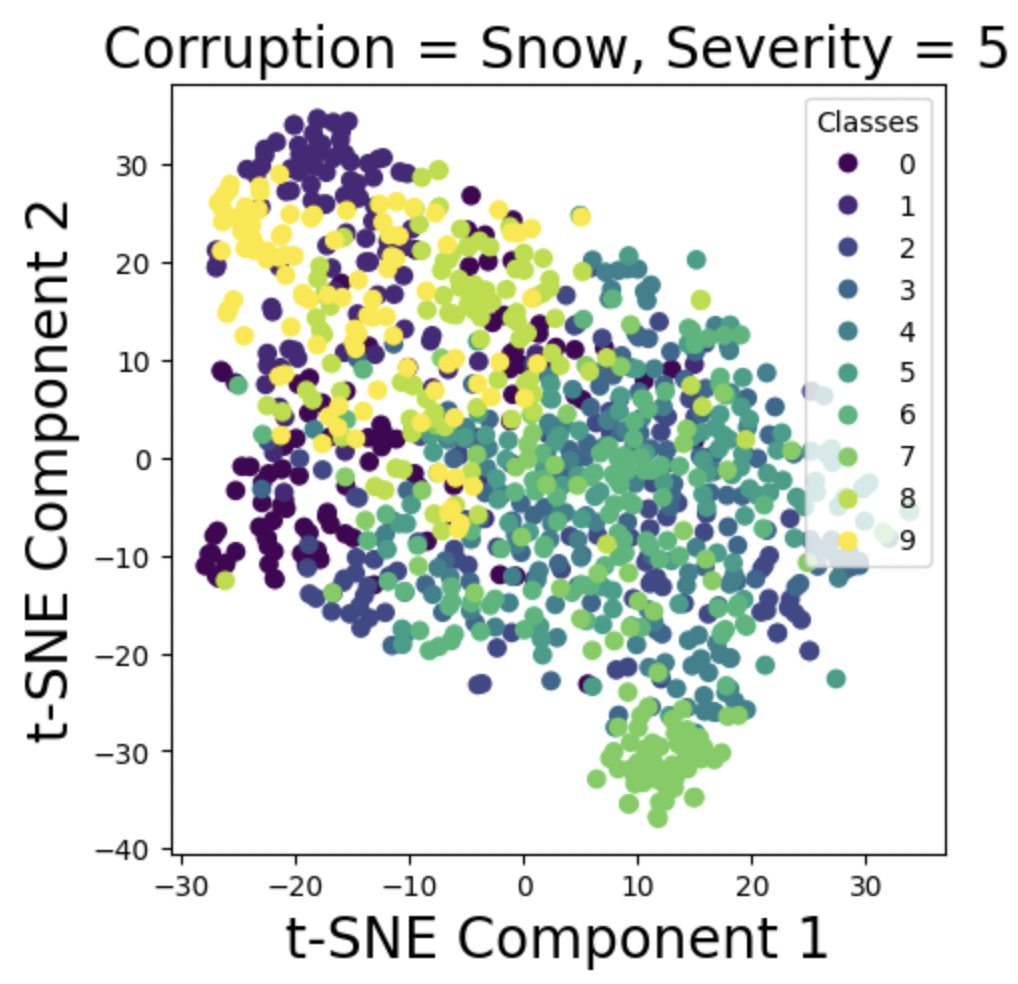} 
    \caption{Feature visualization for CLIP (ViT) with Snow (severity=5)}
  \end{subfigure}
  \vspace{10pt}

  \begin{subfigure}{0.30\textwidth}
    \includegraphics[width=1.2\linewidth]{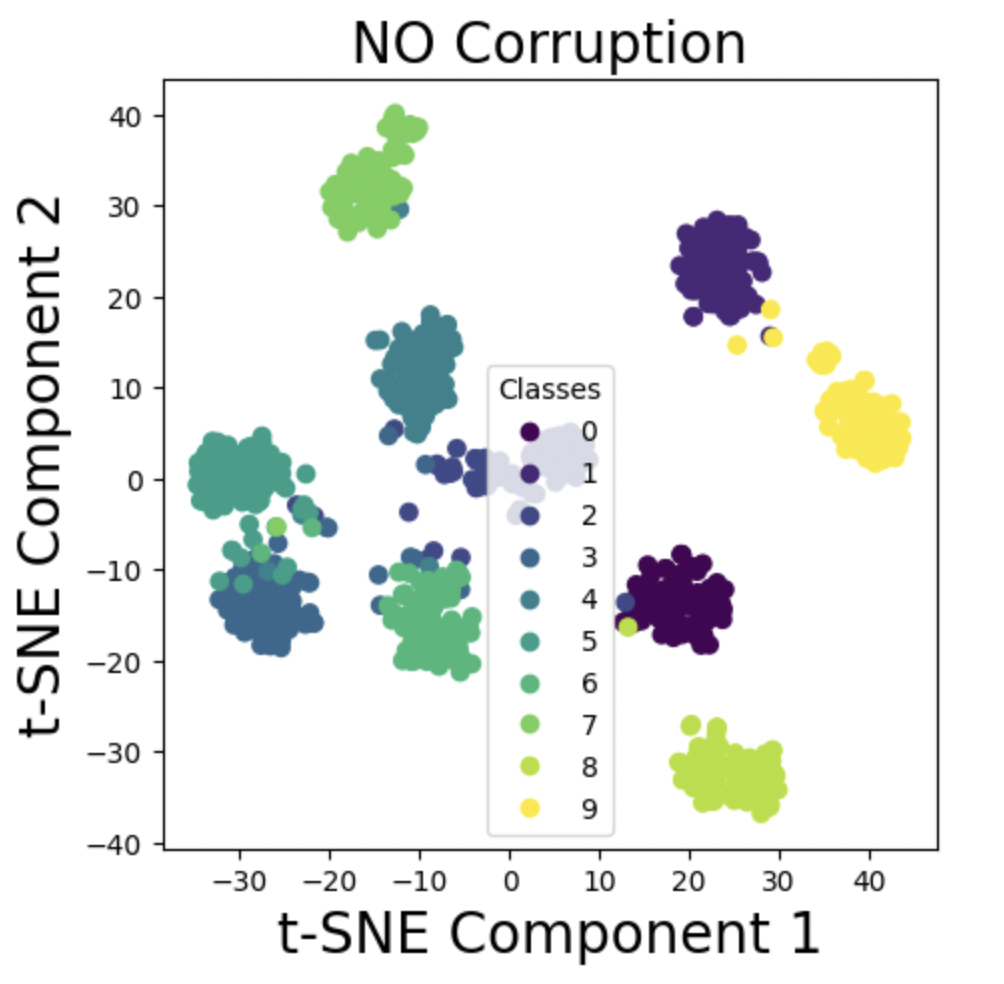} 
    \caption{Feature visualization for CoCa Model with no corruption}
  \end{subfigure}
  \hfill
  \begin{subfigure}{0.30\textwidth}

    \includegraphics[width=1.18\linewidth]{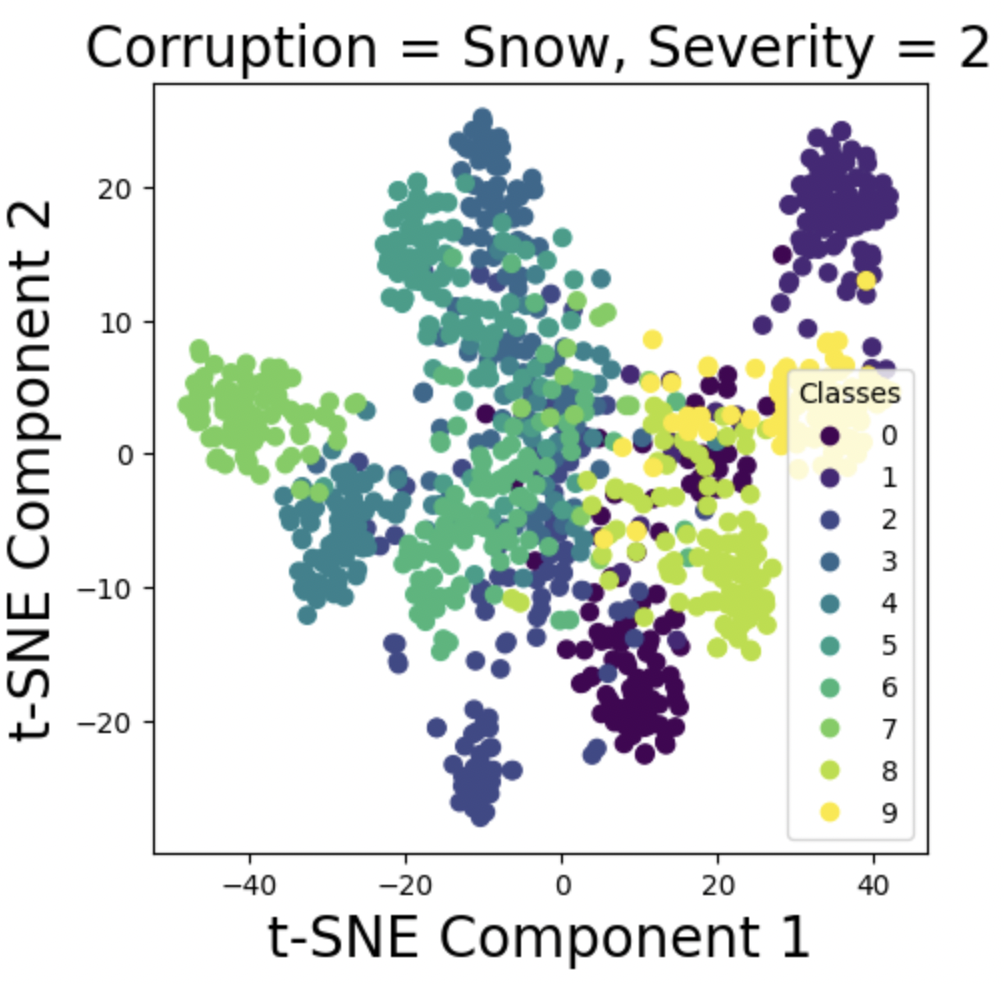} 
    \caption{Feature visualization for CoCa Model with Snow (severity=2)}
  \end{subfigure}
  \hfill
  \begin{subfigure}{0.30\textwidth}
  
    \includegraphics[width=1.2\linewidth]{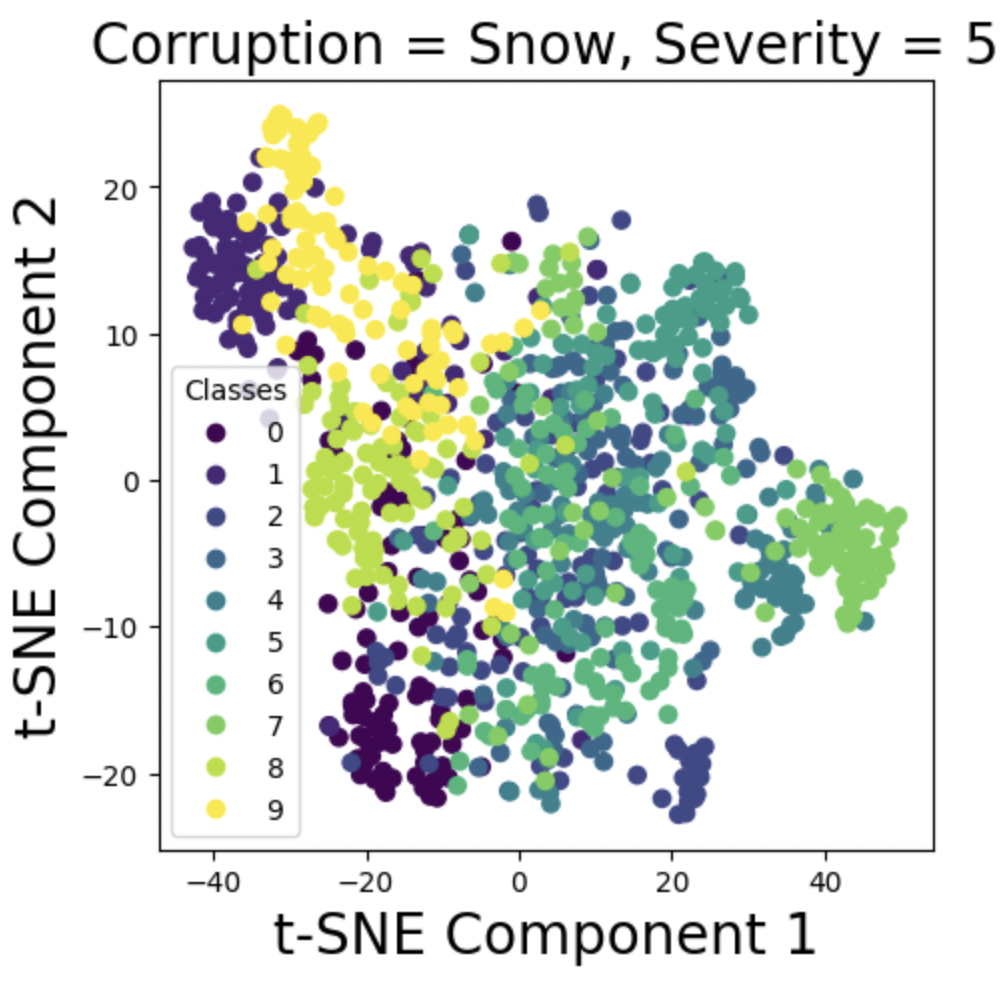} 
    \caption{Feature visualization for CoCa Model with Snow (severity=5)}
  \end{subfigure}

  \caption{Qualitative results for \textbf{CLIP (ViT)} (Contrastive Multi-Encoder model, top row) and \textbf{Coca} (Hybrid Model, bottom row) on \textit{Snow Corruptions} with varying levels of perturbations. We visualize the 512 dimensional latent space embeddings from the visual encoder for CIFAR10 in 2D using t-SNE}
  \label{fig:fullpagefigures}
\end{figure*}
\section*{Conclusion and Findings}
In this work we studied the generalization capability of different types of vision-language foundational models under varying levels of common corruptions. Overall, we find that Hybrid models like Coca and BLIP2 (which use a combination of contrastive and captioning loss) perform better on zero-shot recognition (i.e generalize well to novel classes) and are more robust to common corruptions than Contrastive Multi-Encoder models like CLIP. Their latent representations are stay much more discriminative and semantically meaningful under varying levels of corruptions compared to CLIP counterparts as shown by our latent space analysis. Furthermore, we find that overall ResNet based foundational models are much inferior to vision transformer based models in terms of performance and robustness to corruptions.

As a result of this study we would recommend applied engineers such as those in autonomous driving and robotics industry to use vision transformer based Hybrid models like Coca which learn robust representations due to combination of contrastive (helps global representation) and generative captioning (helps with fine-grained aspects) objectives.



\bibliographystyle{plainnat}
\bibliography{neurips_2020}




\end{document}